# Compresión y análisis de imágenes por medio de algoritmos para la ganadería de precisión.

David Agudelo Tapias
Universidad EAFIT
Colombia
dagudelot@eafit.edu.co

Simón Marín
Universidad Eafit
Colombia
smaring1@eafit.edu.co

Mauricio Toro
Universidad Eafit
Colombia
mtorobe@eafit.edu.co

**RESUMEN**
El problema que se quiere llegar a solucionar en este proyecto de la materia de Estructuras de datos y algoritmos, es llegar a descifrar unas imágenes, las cuales tienen en ellas animales, siendo más específicos, animales vacunos; en las cuales hay que identificar si el animal es sano, o sea, si está en buenas condiciones para luego ser tenido en cuenta en el proceso de selección de la ganadería, o si está enfermo, para saber si se llega a descartar. Todo esto por medio de un algoritmo de compresión, el cual permite tomar las imágenes y llevarlas a una examinación de estas en el código, donde no siempre los resultados van a ser cien por ciento exactos, pero lo que permite a este código ser eficiente, es que se trabaja con aprendizaje de máquina, lo que quiere decir que mientras más información toma, más precisos van a ser los resultados sin traer consigo afectaciones en general.

Los algoritmos propuestos son NN y la interpolación bilineal, donde se obtuivieron resultados significativos sobre la velocidad de ejecución. Se concluye que se pudo haber hecho un mejor trabajo, pero con lo entregado, se cree que es un buen resultado del trabajo.

## 1. INTRODUCCIÓN
La motivación para resolver esta problemática planteada para el proyecto, es llegar a poder comprimir las imágenes de los animales bovinos, para hacer que el código funcione de la manera más eficiente posible y así permitirle saber a las personas que lo usen, qué salud animal tiene cada uno de estos y poder llevar un mejor control, preferiblemente más preciso, de sus ganados en la ganadería.

### 1.1. Problema
El problema que se trabajará es poder identificar por medio de un código y unas imágenes, qué vacunos tiene una salud con el que se puede considerar sano o enfermo cada animal analizado. El impacto que esto genera es que permite que el tratado de estos animales en específico sea el más adecuado para los negocios, ya que al no usarlo o darle un mal uso, se están perdiendo oportunidades en las grandes compañías que trabajan en este sector agrícola.

## 2. TRABAJOS RELACIONADOS

En lo que sigue, explicamos cuatro trabajos relacionados. en el dominio de la clasificación de la salud animal y la compresión de datos. en el contexto del PLF.

### 2.1 Visual Localisation and Individual Identification of Holstein Friesian Cattle via Deep Learning.
Existe una raza de animales vacunos la cual es Holstein Friesian, y esta es representativa para la ganadería por su gran producción de leche. El problema que querían solucionar era identificar por medio de una red de cámaras fijas, y también de vehículos aéreos no tripulados (UAV), el pelaje del animal para saber si era de esta raza tan demandada; estas cámaras tomaban fotos de 3840x2160 pixeles, de una forma rectangular y estando ubicadas a una altura de cinco metros. Usando Long Short-Term Memory se organizan los patrones en las pieles según los animales que haya en el lugar. Los resultados fueron: "Demostramos que la detección y localización del ganado frisón se puede realizar de forma robusta con una precisión del 99,3% en estos datos", "También evaluamos la identificación a través de una tubería de procesamiento de video en 46,430 cuadros provenientes de 34 clips (aproximadamente 20 s de longitud cada uno) de imágenes de UAV tomadas durante el pastoreo (23 individuos, precisión = 98.1%)".

### 2.2 Cloud services integration for farm animals' behavior studies based on smartphones as activity sensors
Para este trabajo, se quería transferir, almacenar, tratar y compartir datos y para esto se usan dos modelos de iPhone, que son los modelos 4s y 5s, ya que los celulares contienen gran cantidad de sensores de todo tipo.

> "Se ejemplifica una aplicación al estudio del comportamiento del ganado en pastos basada en los datos registrados con la IMU de iPhone 4s. También se logra una comparación de rendimiento entre el iPhone 4s y el iPhone 5s. El paquete también viene con una interfaz web para codificar el comportamiento real observado en los videos y sincronizar las observaciones con las señales del sensor. Finalmente, el uso de Edge computing en el iPhone redujo en un 43,5% en promedio el tamaño de los datos sin procesar al eliminar las redundancias. La limitación del número de dígitos en una variable individual puede reducir la redundancia de datos hasta en un 98,5%"

### 2.3 An Animal Welfare Platform for Extensive Livestock Production Systems

El problema a solucionar es acoplarse a las nuevas reglas en la UE para demostrar que los bovinos están en las mejores condiciones, para al final censar que "problemas de salud, estado y bienestar de la granja". La solución que plantearon fue usar un sensor inalámbrico "basado en algoritmos de reconocimiento de patrones de redes neuronales", dando información de sus movimientos, velocidad y donde se encuentran. Los resultados fueron resultados de los patrones, que los agricultores usan para tomar las mejores medidas para sus animales.

## 3. MATERIALES Y MÉTODOS

En esta sección, explicamos cómo se recogieron y procesaron los datos y, después, diferentes alternativas de algoritmos de compresión de imágenes para mejorar la clasificación de la salud animal.

### 3.1 Recopilación y procesamiento de datos

Recogimos datos de *Google Images* y *Bing Images* divididos en dos grupos: ganado sano y ganado enfermo. Para el ganado sano, la cadena de búsqueda era "cow". Para el ganado enfermo, la cadena de búsqueda era "cow + sick".

En el siguiente paso, ambos grupos de imágenes fueron transformadas a escala de grises usando Python OpenCV y fueron transformadas en archivos de valores separados por comas (en inglés, CSV). Los conjuntos de datos estaban equilibrados.

El conjunto de datos se dividió en un 70% para entrenamiento y un 30% para pruebas. Los conjuntos de datos están disponibles en https://github.com/mauriciotoro/ST0245-Eafit/tree/master/proyecto/datasets .

Por último, utilizando el conjunto de datos de entrenamiento, entrenamos una red neuronal convolucional para la clasificación binaria de imágenes utilizando *Teachable Machine* de Google disponible en https://teachablemachine.withgoogle.com/train/image.

### 3.2 Alternativas de compresión de imágenes con pérdida

En lo que sigue, presentamos diferentes algoritmos usados para comprimir imágenes con pérdida.

### 3.2.1 Liquid Rescaling

Es un algoritmo que permite que se agreguen o eliminen píxeles, si es que importan o no respectivamente, ampliando o reduciendo la imagen a analizar.

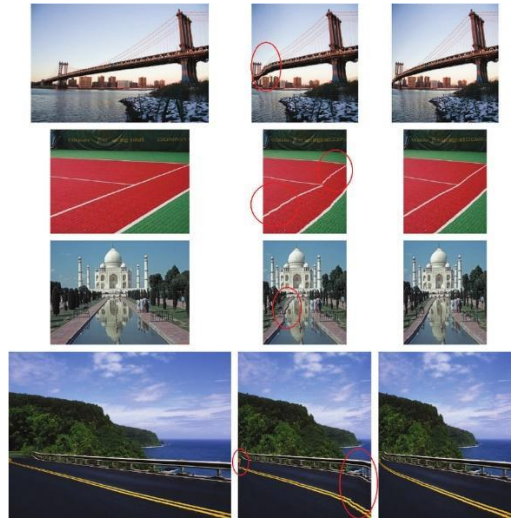

### 3.2.2 Image Scaling

Se trata de un algoritmo que "al momento de escalar un vector gráfico, los elementos primitivos que componen la imagen se pueden modificar usando transformaciones geométricas sin pérdida de la calidad de imagen, cuando se escalan gráficos ráster se debe generar otra imagen con un mayor o menor número de píxeles".

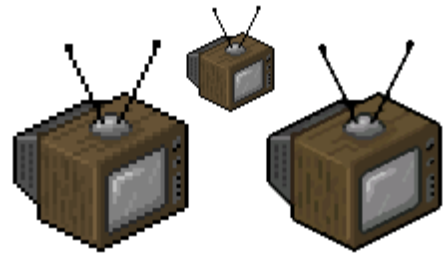

### 3.2.3 Compresión Fractal

Es un algoritmo el cual, por medio de patrones geométricos, la imagen que es analizada pasa a una mejor resolución, y además es óptimo al momento de analizar imágenes de la naturaleza, ya que hay fractales que se repiten.

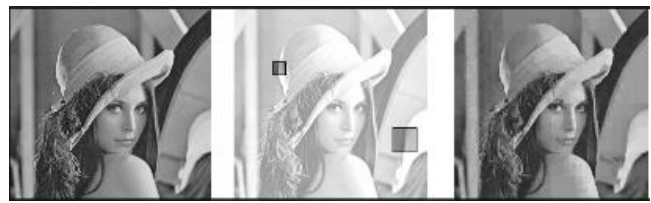

### 3.2.4 Compresión JPEG

Este algoritmo usa los píxeles de una imagen y los agrupa, es poco usado ya que representa una gran pérdida de calidad en la imagen, después de este proceso, y al final de esto se le aplica el algoritmo de Huffman.

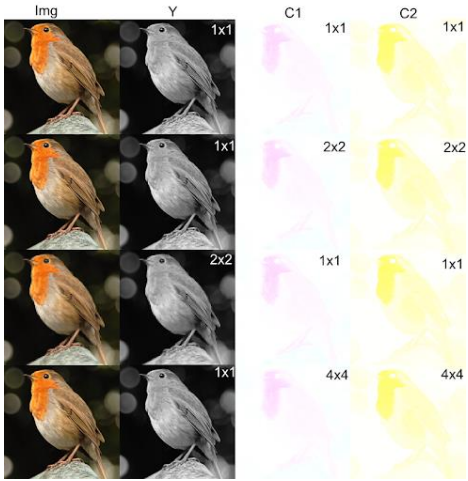

## 3.3 Alternativas de compresión de imágenes sin pérdida

En lo que sigue, presentamos diferentes algoritmos usados para comprimir imágenes sin pérdida.

### 3.3.1 LZ77

Este algoritmo trata de un codificador de diccionario, donde LZ77 "codifica y decodifica desde una ventana deslizante sobre los caracteres vistos anteriormente" y así poder encontrar coincidencias. Su complejidad en memoria es de O(n), pero en el tiempo es más demorado comparado con algunos y su complejidad es $O(n^2)$.

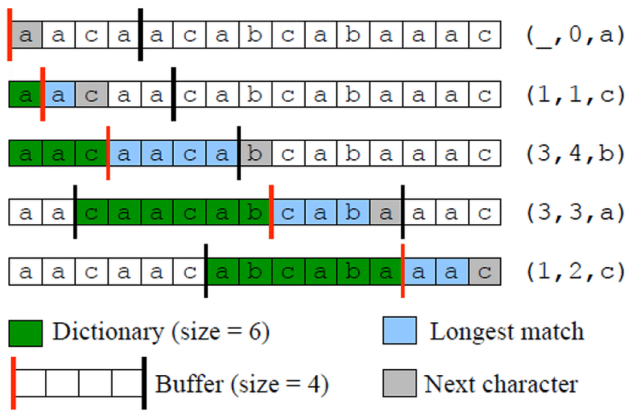

### 3.3.2 Codificación Huffman

Este tipo de algoritmo se trata de un árbol de nodos donde este "permite asignar a los diferentes símbolos a comprimir, un código binario". Su complejidad en memoria es O(n) ya que es lineal, y en el tiempo tiene una complejidad de O(n*Log(n)).

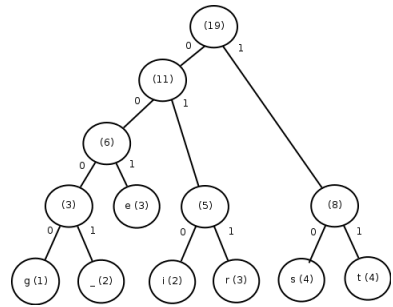

### 3.3.3 Borrow-Wheeler transform

"La transformación de Burrows-Wheeler reorganiza una cadena de caracteres en series similares. Es un algoritmo que prepara los datos para su posterior uso con técnicas de compresión, al ingresar una cadena de caracteres, la transformación conmuta su orden" [2]. Su complejidad en memoria y en tiempo son de O(n).

Figure 2.2: Burrows-Wheeler transform of the string `mississippi`: (a) rotations of the string; (b) sorted matrix; (c) permuted string (last character of sorted matrix); (d) permuted string and sorted string.

### 3.3.4 LZ78

Este algoritmo es una versión mejorada del algoritmo LZ77, donde necesita un diccionario para así, comprimir los datos recibidos. Su complejidad es muy parecida a la de LZ77, "este tiene unas mejoras respecto al consumo de memoria, pero tarda más tiempo".

Figura 9: Ejemplo compresión usando LZ78 de la cadena "DAD DADABDAD"

# 4. DISEÑO E IMPLEMENTACIÓN DE LOS ALGORITMOS

En lo que sigue, explicamos las estructuras de datos y los algoritmos utilizados en este trabajo. Las implementaciones de las estructuras de datos y los algoritmos están disponibles en Github[1].

## 4.1 Estructuras de datos

La estructura de datos que se usó es la matriz, que "es un conjunto ordenado en una estructura de filas y columnas. Los elementos de este conjunto pueden ser objetos matemáticos de muy variados tipos, aunque de forma particular, trabajaremos exclusivamente con matrices formadas por números reales." [8] Cada elemento de la matriz representa el valor del pixel de la imagen a ser analizada y comprimida, donde cada una de estas es dada en un archivo tipo CSV.

**Figura 1:**

```
[0,0,0,0,0,0,...,0,0,0,0,0,0]
[1,0,0,0,1,0,1,...,2,1,1,1,0,0]
[2,1,1,1,1,1,1,...,2,2,2,2,2,2]
[5,3,4,3,3,3,2,...,2,3,3,3,3,4,5]
[4,4,4,2,2,4,3,...,3,2,1,2,2,3,4]
[6,5,5,5,5,5,4,...,2,3,2,2,3,4,5]
[4,4,5,6,6,5,5,...,4,4,5,4,4,4,4]
[3,3,2,4,2,4,3,...,3,2,4,4,4,3,4]
[2,2,3,2,1,2,2,...,4,4,2,3,2,2,2]
[1,2,1,3,1,2,1,...,2,1,2,1,2,3,2]
[0,0,0,0,0,0,...,0,0,0,0,0,0]
```

## 4.2 Algoritmos

En este trabajo, se propone un algoritmo de compresión que es de un algoritmo de compresión de imágenes con pérdida. También explicamos cómo funciona la descompresión para el algoritmo propuesto.

### 4.2.1 Algoritmo de compresión de imágenes con pérdida.

El algoritmo seleccionado para la compresión de imágenes con pérdida fue la interpolación del vecino más cercano.

"Es un método que simplemente busca en las observaciones más cercanas a la que se está tratando de predecir y clasifica el punto de interés basado en la mayoría de datos que le rodean" [7]

"Cuando el método de interpolación del vecino más cercano agranda la imagen, el píxel agregado es el valor del píxel del vecino más cercano. Debido a que el método es simple, la velocidad de procesamiento es muy rápida, pero la calidad de imagen de la imagen ampliada se deteriora significativamente y a menudo contiene bordes irregulares." [8]

Figura:

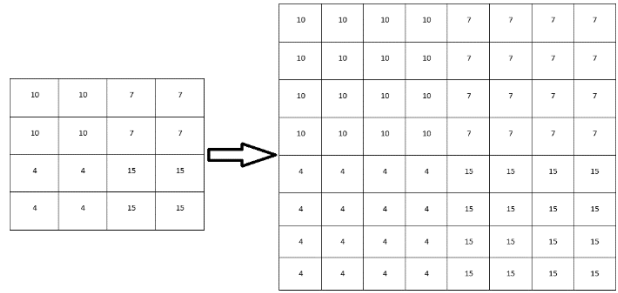

## 4.3 Análisis de la complejidad de los algoritmos

Para el código de compresión se calculó la complejidad fácilmente tanto de tiempo como de memoria, yaque analizando cada línea de este y usando librerías como time para ver los tiempos de ejecución y la librería sys para ver como es el uso de memoria de cada código, se pudo ver que se analizaban las filas y las columnas, y siempre marcaba tiempos muy parecidos para los análisis de cada imagen en diferentes compilaciones.

**Código con pérdida y sin pérdida.**

| Algoritmo | La complejidad del tiempo |
|---|---|
| Compresión | $O(N^{1,75} * M^{1,75})$ |
| Descompresión | $O(N*M)$ |

*Tabla* **2:** N y M en este caso significan las filas y las columnas, y en los análisis hechos para ver los tiempos medidos, se determinó que siempre va a depender de las filas y columnas a analizar.

| Algoritmo | Complejidad de la memoria |
|---|---|
| Compresión | $O(N*M)$ |
| Descompresión | $O(N*M)$ |

---

[1] https://github.com/DavidAgudeloTapias/ST0245-001/tree/master/Proyecto

*Tabla* **3:** N y M en este caso significan las filas y las columnas, y en los análisis hechos para ver los tiempos medidos, se determinó que siempre va a depender de las filas y columnas a analizar.

### 4.4 Criterios de diseño del algoritmo

El algoritmo fue implementado de esta manera, ya que, buscando por un largo tiempo en internet, se encontraron ambos códigos, tanto el que se usó para compresión con pérdidas como el código para compresión sin pérdidas. Se encontraron ambos códigos un mes antes de la entrega final, pero no se sabía si se podía implementar el código de compresión sin perdidas que se usó hasta que el autor Mauricio Toro lo permitió, que fue dos días antes de la entrega final.

## 5. RESULTADOS

### 5.2 Tiempos de ejecución

En lo que sigue explicamos la relación entre el tiempo promedio de ejecución y el tamaño promedio de las imágenes del conjunto de datos completo, en la Tabla 6.

|  | *Tiempo promedio de ejecución (s)* | *Tamaño promedio del archivo (MB)* |
|---|---|---|
| *Compresión* | 5.7 s | 20.7 MB |
| *Descompresión* | 2.2 s | 20.7 MB |

**Tabla 6:** Tiempo de ejecución de los algoritmos *(Por favor, escriba el nombre de los algoritmos, por ejemplo, tallado de costuras y LZ77)* para diferentes imágenes en el conjunto de datos.

### 5.3 Consumo de memoria

Presentamos el consumo de memoria de los algoritmos de compresión y descompresión en la Tabla 7.

|  | *Consumo promedio de memoria (MB)* | *Tamaño promedio del archivo (MB)* |
|---|---|---|
| Compresión | 142.3 MB | 20.7 MB |
| Descompresión | 141.1 MB | 20.7 MB |

**Tabla 7:** Consumo promedio de memoria de todas las imágenes del conjunto de datos, tanto para la compresión como para la descompresión.

### 5.3 Tasa de compresión

Presentamos los resultados de la tasa de compresión del algoritmo en la Tabla 8.

|  | *Ganado sano* | *Ganado enfermo* |
|---|---|---|
| Tasa de compresión promedio | 2.3 : 1 | 2.3 : 1 |

**Tabla 8:** Promedio redondeado de la tasa de compresión de todas las imágenes de ganado sano y ganado enfermo.

## 6. DISCUSIÓN DE LOS RESULTADOS

Los resultados obtenidos haciendo este proyecto se consideran buenos, ya que la velocidad de ejecución del programa es rápida a comparación de otros códigos, aunque si está sobreajustado ya que se adaptó un código con pérdidas a uno sin pérdidas, y el consumo de memoria no es tan ato como se esperaba. Como no se pudo lograr el objetivo final de este proyecto que era saber si la imagen analizada era sobre un ganado enfermo o sano, no se lograron probar lo datos de prueba y   obtener la exactitud del programa.

### 6.1 Trabajos futuros

A David Agudelo le gustaría mejorar en su disposición para trabajar ya que este proyecto lo realizó solo por no conocer a nadie del grupo, así que se compromete a conocer gente y realizar mejores entregas. Viendo hacia el futuro, el código podría mejorar de tal manera que se halle un código de compresión sin pérdidas como los sugeridos.

### RECONOCIMIENTOS

## Referencias.

1). Wikipedia. *LZ77 and LZ78*. Wikipedia: la enciclopedia libre. Retrieved Agosto 14, 2021, from https://en.wikipedia.org/wiki/LZ77_and_LZ78

2). Castro, J. E., Vélez, C. G., & Mesa, J. E. *Algoritmos de compresión para la optimización del consumo de baterías en ganadería de precisión*. Retrieved August 15, 2021, from https://osf.io

3). Fernández, S. *Código Huffman*. Retrieved August 14, 2021, from https://es.slideshare.net/mejiaff/cdigo-huffman

4). Debauche, O. *Cloud services integration for farm animals' behavior studies based on smartphones as activity sensors*. Retrieved August 16, 2021, from https://link.springer.com/article/10.1007/s12652-018-0845-9

5). Andrew, W., Greatwood, C., & Burghardt, T. *Visual Localisation and Individual Identification of Holstein Friesian Cattle via Deep Learning*. Retrieved August 16, 2021, from https://openaccess.thecvf.com/content_ICCV_2017_workshops/w41/html/Andrew_Visual_Localisation_and_ICCV_2017_paper.html

6). Doulgerakis, V., Kalyvas, D., Bocaj, E., Giannousis, C., Feidakis, M., Laliotis, G. P., Patrikakis, C., & Bizelis, I. *An Animal Welfare Platform for Extensive Livestock Production Systems*. ResearchGate. Retrieved August 16, 2021, from https://www.researchgate.net/publication/338595895_An_Animal_Welfare_Platform_for_Extensive_Livestock_Production_Systems

7). Bagnato, J. I. *Clasificar con K-Nearest-Neighbor ejemplo en Python*. Retrieved October 10, 2021, from https://www.aprendemachinelearning.com/clasificar-con-k-nearest-neighbor-ejemplo-en-python/#:~:text=K%2DNearest%2DNeighbor%20es%20un,el%20mundo%20del%20Aprendizaje%20Autom%C3%A1tico.

8) *Python implementa la interpolación bilineal, la interpolación del vecino más cercano y la interpolación cúbica*. Retrieved October 16, 2021, from http://recursostic.educacion.es/descartes/web/materiales_didacticos/Calculo_matricial_d3/defmat.htm